\documentclass{article}

\usepackage{arxiv}

\usepackage[utf8]{inputenc} 
\usepackage[T1]{fontenc}    
\usepackage{hyperref}       
\usepackage{url}            
\usepackage{booktabs}       
\usepackage{amsfonts}       
\usepackage{nicefrac}       
\usepackage{microtype}      
\usepackage{cleveref}       
\usepackage{lipsum}         
\usepackage{graphicx}
\usepackage{natbib}
\usepackage{doi}

\usepackage{amssymb}
\usepackage{makecell}
\usepackage{enumitem}
\setitemize{noitemsep,topsep=0pt,parsep=0pt,partopsep=0pt,leftmargin=*}
\usepackage[framemethod=tikz]{mdframed}
\usepackage[ampersand]{easylist}
\usepackage{graphicx}

\usepackage{caption}

\title{Textual Summarisation of Large Sets: Towards a General Approach}

\newif\ifuniqueAffiliation
\uniqueAffiliationtrue

\ifuniqueAffiliation 

\author{
    Kittipitch Kuptavanich\\
    Department of Computing Science\\
    University of Aberdeen\\
    Aberdeen, UK \\
    {\tt kittipitch.kuptavanich@abdn.ac.uk} \And
    Ehud Reiter \\
    Department of Computing Science\\
    University of Aberdeen\\
    Aberdeen, UK \\
    {\tt e.reiter@abdn.ac.uk}\AND
    Kees Van Deemter \\
    Utrecht University\\
    Utrecht, NL\\
    {\tt k.vandeemter@uu.nl} \And
    Advaith Siddharthan \\
    Knowledge Media Institute\\
    The Open University\\
    Milton Keynes, UK \\
    {\tt advaith.siddharthan@open.ac.uk}
}

\date{}
\else
\usepackage{authblk}

\newbox{\orcid}\sbox{\orcid}{\includegraphics[scale=0.06]{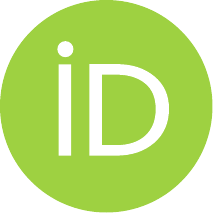}}
\author[1]{%
    \href{https://orcid.org/0000-0000-0000-0000}{\usebox{\orcid}\hspace{1mm}David S.~Hippocampus\thanks{\texttt{hippo@cs.cranberry-lemon.edu}}}%
}
\author[1,2]{%
    \href{https://orcid.org/0000-0000-0000-0000}{\usebox{\orcid}\hspace{1mm}Elias D.~Striatum\thanks{\texttt{stariate@ee.mount-sheikh.edu}}}%
}
\affil[1]{Department of Computer Science, Cranberry-Lemon University, Pittsburgh, PA 15213}
\affil[2]{Department of Electrical Engineering, Mount-Sheikh University, Santa Narimana, Levand}
\fi

\renewcommand{\shorttitle}{\textit{arXiv} Template}

\hypersetup{
pdftitle={A template for the arxiv style},
pdfsubject={q-bio.NC, q-bio.QM},
pdfauthor={David S.~Hippocampus, Elias D.~Striatum},
pdfkeywords={First keyword, Second keyword, More},
}

\begin{document}
\maketitle
\begin{abstract}
We are developing techniques to generate summary descriptions of sets of objects.  In this paper, we present and evaluate a rule-based NLG technique for summarising sets of bibliographical references in academic papers. This extends our previous work on summarising sets of consumer products and shows how our model generalises across these two very different domains.
\end{abstract}

\section{Introduction}

Shneiderman's mantra, \textit{``Overview first, zoom and filter, then
details-on-demand''}, highlights the importance of giving readers a high-level overview before offering detail. We apply this idea to generate an overview of sets of objects, hypothesising that an overview will be beneficial to readers who want to understand the set.
Previously we investigated the domain of consumer products, focusing on descriptions of  products (such as TVs) which are intended to help readers decide which specific products to buy. Now we aim to generalise the techniques we have developed, by looking at a very different type of domain, namely bibliographical references in academic papers.

\section{Related Work}
\label{sec:related_work}
\subsection{Text summarising relating to sets of items}

There exists many works on summarisation \citep{nenkova2012survey, gambhir2017recent}, but they usually take text as input. In many situations, however, a computer system has structured (i.e., non-textual) information about a large set of entities, where it is desirable that human users can obtain a good understanding of what’s in the set. For instance, for a consumer who wants to buy a TV online, there are many websites that provide product comparison interfaces\footnote{gadgetsnow.com, saveonlaptops.com, pricespy.co.uk, opodo.com, uswitch.com, www.amazon.com} in a tabular format and also reviews of products written by hands\footnote{which.co.uk, www.consumerreports.org}. There are works related to describing sets of objects, for example, \citep{Deemter2002} for generating noun phrase to identify intended sets, \citep{kutlak2016production} in trying to identify interestingness, identifying surprisingness among different attributes \citep{Geng:2006:IMD:1132960.1132963}, for descriptions of products \citep{weissgraeber2017working}, to generate a page title for set of items with shared properties from metadata \citep{mathur2017generating}, or to generate titles for web tables \citep{hancock2019generating}. However, we are not aware of a system that describes/summarises a set of items from collections of data.

\subsection{Summarising sets of consumer products}
\label{subsec:prodSet}

In previous work, we took inspiration from handwritten summaries and describe a set of TVs, for example, from a search result, by relating it to its superset. The idea is to produce sentences like: ``32 inch TVs are generally less expensive''.

This algorithm, which we will now call [prodSet], gave rise to summaries with this structure:
\paragraph{Introductory paragraph,} where we report the shape of the price curve by providing its range and centering.
\paragraph{Listing of products' important features,} using a variety of quantifiers, i.e., \textit{most}, \textit{a large portion}, and \textit{some} (e.g., ``\textit{some} TVs have built-in WiFi'') and making vague comparisons to its superset (e.g., ``TVs with a Smart-Internet feature are \textit{generally slightly more expensive (\textsterling475 vs \textsterling450)}'')

\subsection{Lessons from our previous work}
We tentatively extracted the following lessons:
\begin{itemize}
    \item Feature selection can be informed by (1) statistical analysis of the database (with a known dominating column), (2) corpus analysis or (3) surveys of users.
    \item When there is a dominating column (e.g., price), other features should be related to this column, for example, features that have stronger influence on prices are more important to the readers.
    \item Important continuous variables should be reported using range and median
    \item Important non-continuous features should be described using quantifiers.
\end{itemize}
\section{Summarising sets of references}
We want to put these ideas to the test by looking at a domain that is very different from the one we studied before. We chose literature references in academic papers, for various reasons. Firstly, we are very familiar with this domain. Secondly, there is no obvious dominating attribute, like the ``price'' column we had with consumer products (a good start toward generalisation). And lastly, the algorithm could be used in other cases, e.g., for characterising an author (by summarising her publications), or a scientific journal (by summarising the papers that have appeared in it).

\subsection{A Survey of reader preferences}
When there is no suitable corpus, as stated earlier in our lessons learned, we can obtain information to perform feature selections using other means. In this case, we decided to conduct a survey of readers'  preferences, to investigate how researchers read the references of a paper, and what they hope to learn in different scenarios. The scenarios of interest were: reviewing (for a conference or a journal), learning the content (algorithm, findings, etc), writing another paper (hence need to look for citation materials), or browsing (and deciding whether a paper is worth reading). The Participants were graduate 45 students/researchers from different discipline recruited by the college of the lead author's mailing list.

\paragraph*{Lessons for a summarisation algorithm:}
We cannot give detailed results from our survey because of space limitations, but its key findings were: (1) There is an implicit dominating column, namely the citation counts. (2) The readers want to know the prominent authors in the field. (3) The readers want to know whether prominent works are cited. (4) The important attributes in references are venue type, year, self-citation and the academic domain and sub-domain of the venue.

Lesson (1) means that a summarisation algorithm for this domain can work along very similar lines to the one for consumer products (Section~\ref{subsec:prodSet}). To take lessons (2) and (3) onboard, we added information about prominent works and authors. Additionally, we added numbers where initially there were only vague quantifying words.

\subsection{Algorithm}
\label{subsec:algoritm}

\begin{figure*}[ht]
    \includegraphics[width=\textwidth]{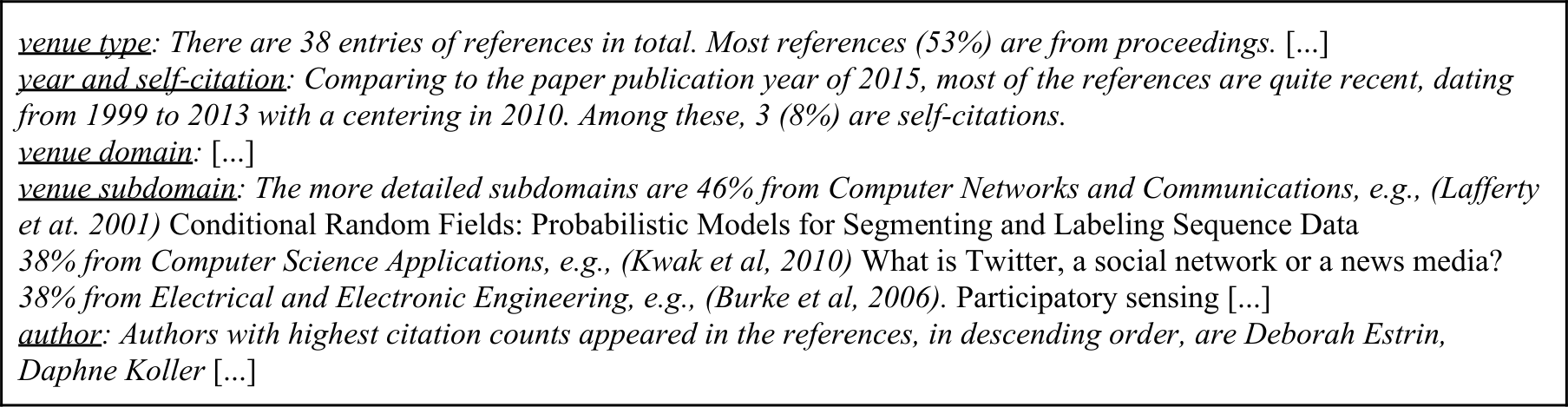}
    \caption{An excerpt from a Summary Generated by [refSet]}
    \label{fig:refSet_summary_example}
\end{figure*}

From the analysis of the survey, we describe here the new algorithm [refSet] used in describing the references.
\paragraph*{(a) Shape of the dominating attribute:} Although we learned that the ``citation count'' is an implicit dominating attribute, readers did not show interest in this attribute, so we decided that we would not report it.
That left the ``introductory sentences'' to be just the number of references in total. We combine this to the description of the first important attribute which is the ``venue type'' (the first paragraph in Figure~\ref{fig:refSet_summary_example}).

\paragraph*{(b) Quantifiers:} Here we try to deliver the sense of clustering of different properties in an attribute. So we generate sentences with quantifying words, i.e., ``most'', ``a large portion'' and ``some'' and the percentages. For example, when describing the venue types, we have ``\textit{Most references (55\%) are from proceedings. A large proportion is from journals (30\%). Some are from books (15\%)}''.
In case of continuous numbers, like \textit{years}, \textit{lengths} or \textit{weights}, we instead, report the range and centering. Also since the description of ``self-citation'' attribute is short, we combined it with the description of the ``year'' attribute.

\paragraph*{(c) Venue subdomains' top publications:} Here, in addition to the normal description of the venue subdomains, for each subdomain, we present a publication from the references with the highest citation count.

\paragraph*{(d) A list of authors with highest citation count:} Lastly, we list 7 authors with the highest citation counts (the last paragraph of Figure~\ref{fig:refSet_summary_example}).

\section{Evaluation}
In this experiment, we used human rating evaluation like we did in our previous experiments.
\subsection{Method}
\paragraph*{Materials:} We selected three papers from 2 Computing Science journals, published between 2014 - 2015, that cite different number of papers, i.e., \citet{marimon2014automatic} (20 references), \citet{anantharam2015extracting} (43 references), and \citet{wang2015peacock} (57 references).

\begin{figure}[!ht]
    \hspace{0pt}\includegraphics[width=0.475\textwidth]{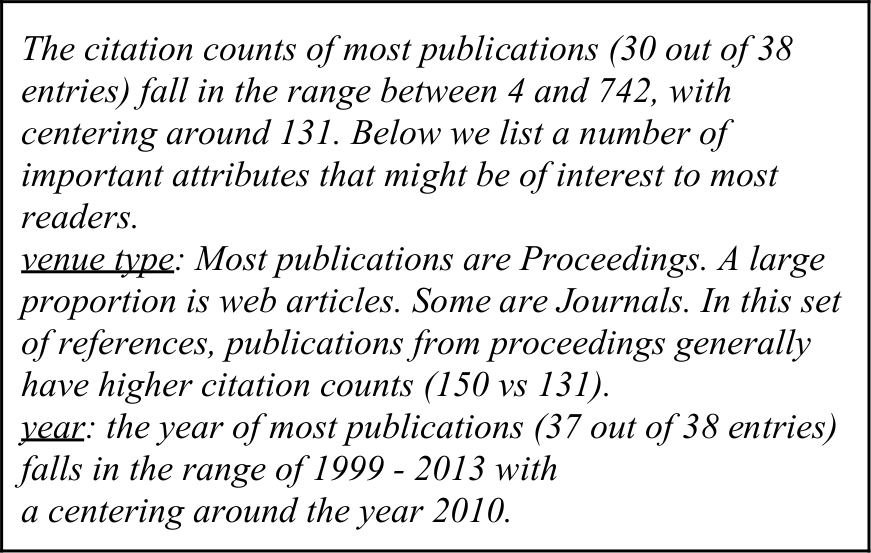}
    \caption{An Excerpt from a Summary Generated by [prodSet]}
    \label{fig:prodSet_example}
\end{figure}
For each publication we used two algorithms, [prodSet] (substituting citation count for price) as shown in Figure~\ref{fig:prodSet_example} and [refSet] (Figure~\ref{fig:refSet_summary_example}), to produce two different summaries  . As in our experiments, the process of document structuring through realisation was carried out using schemata \citep{McKeown1985} approach. The summaries were generated using Jinja2\footnote{jinja.pocoo.org} template engine library. We had 2 cases of the baselines. The first baseline was where there was no summary provided [nosum]. And another baseline is the reference overview generated from \textit{Semantic Scholar} [sem]\footnote{https://www.semanticscholar.org/}. As shown in Figure~\ref{fig:sem_example}, \textit{Semantic Scholar} displays references sorted in a descending order of influence. We only displayed up to 10 items in the lists provided.
\paragraph*{Participants:} Participants were 28 graduate students/researchers recruited by the departmental mailing list and also social media, e.g., Facebook and twitter.

\paragraph*{Design and Procedure:}
For each of the 3 publications, the participants were presented with the names and sources of the publication, the list of its references along with summaries generated by [prodSet], [refSet] and [sem]. Afterwards, they were asked to rank the choices of summary A, B, C and having no summary at all [nosum] in order of usefulness (\#1 being most useful) under two different scenarios:

\begin{figure}[ht]
    \hspace{2pt}\includegraphics[width=0.475\textwidth]{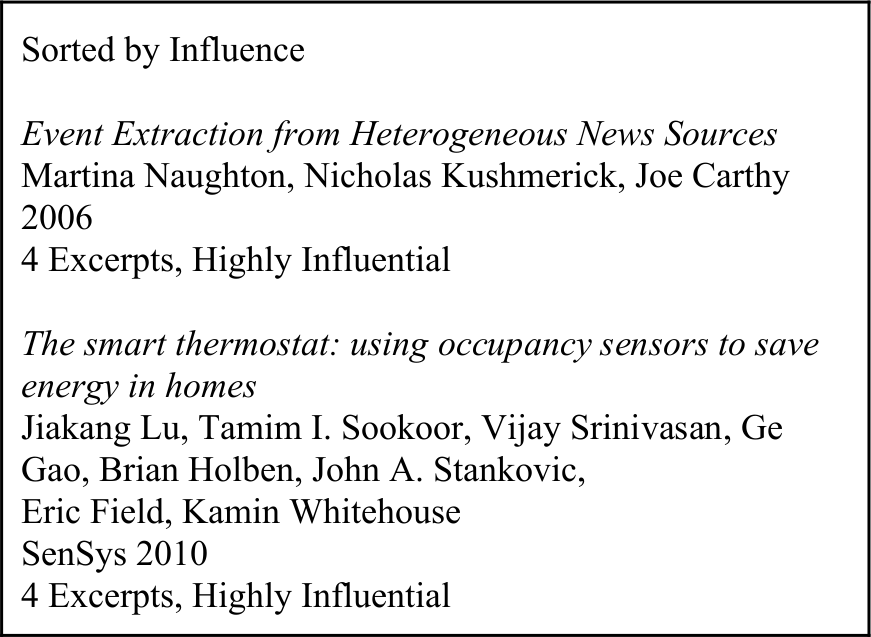}
    \caption{An excerpt from a summary taken from [sem]}
    \label{fig:sem_example}
\end{figure}
b
\begin{itemize}
    \item Scenario 1: Reading for reviewing for a conference or a journal (whether to accept it), writing another paper (whether to cite it), or just browsing (whether to read it further).
    \item Scenario 2: Reading to learn the content, e.g., algorithm, findings, etc.
\end{itemize}
Note that the first scenario is a grouping of 3  scenarios from the survey conducted in the previous section, for they have similar propose of deciding whether to select the paper being read. In each ranking, no ties were allowed.

Then they were asked 6 free text questions:
\begin{enumerate}[label={}]
    \addtolength{\itemindent}{-20pt}
    \vspace{-5pt}\item {[}Q1{]}-{[}Q3{]}: ``Which part of summary [A/B/C] do you think is most useful? (please cut and paste)''
    \vspace{-5pt}\item {[}Q4{]}-{[}Q6{]}: ``What would you like to see added, removed, or changed in the summary [A/B/C]?''
\end{enumerate}


\paragraph*{Hypotheses:}
Our research hypotheses were:
\begin{enumerate}[label={}]
    \addtolength{\itemindent}{-20pt}
    \vspace{-5pt}\item {[}Hyp1{]}: Participants prefer [refSet] summary over having no summary [nosum]
    \vspace{-5pt}\item {[}Hyp2{]}: Participants prefer [refSet] summary over [sem] summary
    \vspace{-5pt}\item {[}Hyp3{]}: Participants prefer [refSet] summary over [prodSet] summary

\end{enumerate}

\subsection{Results}

\paragraph*{Summary Preference:}

\begin{figure}[ht]
    \hspace{-8pt}\includegraphics[width=0.52\textwidth]{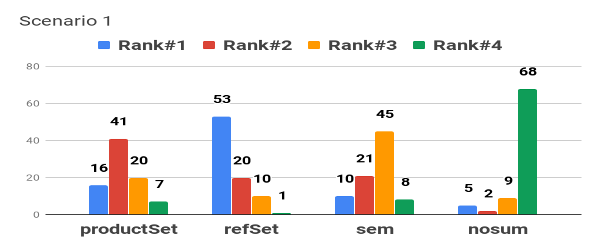}
    \caption{Ranking Counts in Scenario 1}
    \label{fig:ranking_bar_chart1}
\end{figure}

\begin{figure}[ht]
    \hspace{-8pt}\includegraphics[width=0.52\textwidth]{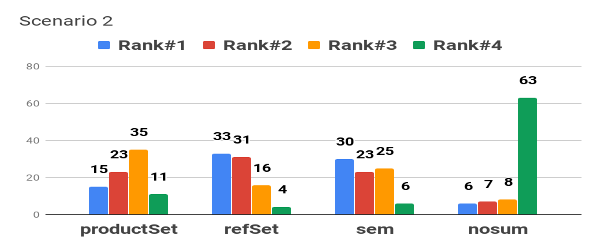}
    \caption{Ranking Counts in Scenario 2}
    \label{fig:ranking_bar_chart2}
\end{figure}

The frequencies of participant rankings for each summary, together with rankings for no summary, for scenario 1 and 2, are shown in Figure~\ref{fig:ranking_bar_chart1} and ~\ref{fig:ranking_bar_chart2} respectively.
We performed an ANOVA analysis \citep{field2012discovering} on the ranking data in each scenario separately.
The result shows that the \textit{number of references} in a paper has no significant effect on the ranking.

In scenario 1, the ANOVA test shows that the \textit{summaries} have significant effects on the ranking (p-value $<2\times10^{-16}$).
A posthoc Tukey HSD \citep{tukey1949comparing}, as shown in Table~\ref{table:tukey_hsd1}, showed that there were significant difference between each summary (p-value $=0.000$), and thus, all [Hyp1], [Hyp2], and [Hyp3] are confirmed.

\begin{table}[ht]
    \centering
    \begin{tabular}{crr}
        \Xhline{2\arrayrulewidth}
            comparison     & diff avg ranking          &  p-value \\
        \hline
        prodSet-refSet     & 0.702 &   0.000   \\
        sem-refSet      & 1.095 &   0.000   \\
        nosum-refSet    & 2.155 &   0.000  \\
        \Xhline{2\arrayrulewidth}
    \end{tabular}
    \caption{A posthoc Tukey HSD on Scenario 1} 
    \label{table:tukey_hsd1} 
\end{table}

In scenario 2, an ANOVA test also shows that the summaries have significant effects on the ranking also with the p-value $<2 \times 10^{-16}$.
However, as shown in Table~\ref{table:tukey_hsd2}, although the mean ranking of [refSet] is better than [sem], the difference is not statistically significant (p-value $=0.548$). So only [Hyp1] and [Hyp2] are confirmed with p-values of $0.000$ in both cases.

\begin{table}[ht]
    \centering
    \begin{tabular}{crr}
        \Xhline{2\arrayrulewidth}
            comparison     & diff avg ranking          &  p-value \\
        \hline
            prodSet-refSet & 0.607 & 0.000\\
            sem-refSet & 0.190 & 0.548\\
            nosum-refSet & 1.631 & 0.000\\
        \Xhline{2\arrayrulewidth}
    \end{tabular}
    \caption{A posthoc Tukey HSD on Scenario 2} 
    \label{table:tukey_hsd2} 
\end{table}
\vspace{-2mm}
\paragraph*{Free text answers:} From the free text answer, what the readers like the most in [refSet] in descending order are passages on venue subdomain, authors, venue domain, year and venue type and self-citation. Some readers would like to see keywords added to the summary.
There are a number of comments wanting to see features of [refSet] and [sem] combined.


\section{Conclusion}
In the first scenario, the [refSet] summary received the best ranking average distinguishedly. In the second scenario, although we cannot say that [refSet] is significantly better than the [sem] summary, its ranking average is still the best. This tells us that our algorithm performs well in helping readers make a decision about a paper, but is less powerful, although preferred by many, in helping them understand its contents. One of the reasons could be that to help readers understand a paper, we need to look at its contents, as the [sem] algorithm does. However, that is beyond our scope of analysing lists of references and trying to give readers overviews of those lists.

On the big picture, we can say that our attempt to generalise our algorithm, derived from the consumer product domain, has a good performance with an encouraging result.
However,  there still needs to further our exploration towards generalisation. For example, in cases where no domain knowledge is readily available, e.g., no present suitable corpus and the situation not permitting a survey, we need to improve our algorithm to be able to work using data analysis alone, e.g., predicting dominating columns by statistical analysis.





\bibliography{acl2019,references_kitt,websites_kitt}

\end{document}